\documentclass[letterpaper, 10 pt, conference]{IEEEtran}
\IEEEoverridecommandlockouts

\usepackage[letterpaper,left=.75in,right=.75in,top=.75in,bottom=.75in]{geometry}

\usepackage{epsfig}
\usepackage{amsmath}
\usepackage{amssymb}
\usepackage{multirow}
\usepackage{graphicx}
\usepackage{tabularx}
\usepackage{booktabs}
\usepackage{arydshln}
\usepackage{color}
\usepackage{subfig}
\usepackage{tikz}
\usepackage[pagebackref=true,breaklinks=true,letterpaper=true,colorlinks,bookmarks=false]{hyperref}
\definecolor{darkgreen}{rgb}{0, 0.7, 0}

\newcommand\copyrighttext{
\centering \footnotesize \copyright 2021 IEEE. Personal use of this material is permitted. Permission from IEEE must be obtained for all other uses, in any current or future media, including reprinting/republishing this material for advertising or promotional purposes, creating new collective works, for resale or redistribution to servers or lists, or reuse of any copyrighted component of this work in other works.}
\newcommand\copyrightnotice{
\begin{tikzpicture}[remember picture,overlay]
\node[anchor=south,yshift=10pt] at (current page.south) {\fbox{\parbox{\dimexpr\textwidth-\fboxsep-\fboxrule\relax}{\copyrighttext}}};
\end{tikzpicture}
}

\definecolor{unlabeled}{rgb}{0.0, 0.0, 0.0}
\definecolor{car}{rgb}{0.39215686274509803, 0.5882352941176471, 0.9607843137254902}
\definecolor{bicycle}{rgb}{0.39215686274509803, 0.9019607843137255, 0.9607843137254902}
\definecolor{motorcycle}{rgb}{0.11764705882352941, 0.23529411764705882, 0.5882352941176471}
\definecolor{truck}{rgb}{0.3137254901960784, 0.11764705882352941, 0.7058823529411765}
\definecolor{othervehicle}{rgb}{0.0, 0.0, 1.0}
\definecolor{person}{rgb}{1.0, 0.11764705882352941, 0.11764705882352941}
\definecolor{bicyclist}{rgb}{1.0, 0.1568627450980392, 0.7843137254901961}
\definecolor{motorcyclist}{rgb}{0.5882352941176471, 0.11764705882352941, 0.35294117647058826}
\definecolor{road}{rgb}{1.0, 0.0, 1.0}
\definecolor{parking}{rgb}{1.0, 0.5882352941176471, 1.0}
\definecolor{sidewalk}{rgb}{0.29411764705882354, 0.0, 0.29411764705882354}
\definecolor{otherground}{rgb}{0.6862745098039216, 0.0, 0.29411764705882354}
\definecolor{building}{rgb}{1.0, 0.7843137254901961, 0.0}
\definecolor{fence}{rgb}{1.0, 0.47058823529411764, 0.19607843137254902}
\definecolor{vegetation}{rgb}{0.0, 0.6862745098039216, 0.0}
\definecolor{trunk}{rgb}{0.5294117647058824, 0.23529411764705882, 0.0}
\definecolor{terrain}{rgb}{0.5882352941176471, 0.9411764705882353, 0.3137254901960784}
\definecolor{pole}{rgb}{1.0, 0.9411764705882353, 0.5882352941176471}
\definecolor{trafficsign}{rgb}{1.0, 0.0, 0.0}

\newcommand\semcolor[1][black]{\textcolor{#1}{\rule{2.2mm}{2.2mm}}}

\title{\LARGE PillarSegNet: Pillar-based Semantic Grid Map Estimation using Sparse LiDAR Data}
\author{Juncong Fei$^{1,2}$, Kunyu Peng$^{2}$, Philipp Heidenreich$^{1}$, Frank Bieder$^{2}$, and Christoph Stiller$^{2}$
\thanks{$^{1}$\,Stellantis, Opel Automobile GmbH, 65423 R\"usselsheim am Main, Germany. Corresponding author email: juncong.fei@gmail.com}
\thanks{$^{2}$\,Institute of Measurement and Control Systems, Karlsruhe Institute of Technology (KIT), 76131 Karlsruhe, Germany.}
}

\begin{document}
\maketitle
\thispagestyle{empty}
\pagestyle{empty}

\copyrightnotice
\begin{abstract}
Semantic understanding of the surrounding environment is essential for automated vehicles.
The recent publication of the \mbox{SemanticKITTI} dataset stimulates the research on semantic segmentation of LiDAR point clouds in urban scenarios.
While most existing approaches predict sparse point-wise semantic classes for the sparse input LiDAR scan, we propose \mbox{PillarSegNet} to be able to output a dense semantic grid map.
In contrast to a previously proposed grid map method, \mbox{PillarSegNet} uses \mbox{PointNet} to learn features directly from the 3D point cloud and then conducts 2D semantic segmentation in the top view.
To train and evaluate our approach, we use both sparse and dense ground truth, where the dense ground truth is obtained from multiple superimposed scans.
Experimental results on the \mbox{SemanticKITTI} dataset show that \mbox{PillarSegNet} achieves a performance gain of about 10\% mIoU over the \mbox{state-of-the-art} grid map method. 
\end{abstract}
\section{Introduction}
\label{sec:introduction}
Understanding the surroundings perceived by multiple onboarding sensors is crucial in many autonomous systems such as automated vehicles. 
To achieve reliable scene understanding, automated vehicles are typically equipped with complementary sensors such as cameras, LiDARs, and radars. 
As one of the key tasks in scene understanding, semantic segmentation associates each pixel of an image or point in a LiDAR point cloud with a semantic class.
While image-based semantic segmentation has been well studied, less research in semantic segmentation of LiDAR point clouds has been conducted due to the lack of publicly available datasets for this task. 
To close this gap, \mbox{SemanticKITTI}~\cite{behley2019iccv} has recently been published as the first large-scale dataset for semantic scene understanding using LiDAR.
In contrast to conventional camera images, LiDAR point clouds provide more precise distance measurements of the 3D world and preserve the geometric information of objects, empowering a better understanding of the 3D surroundings.

\begin{figure}[t] 
\begin{center}
\includegraphics[width=1.0\columnwidth]{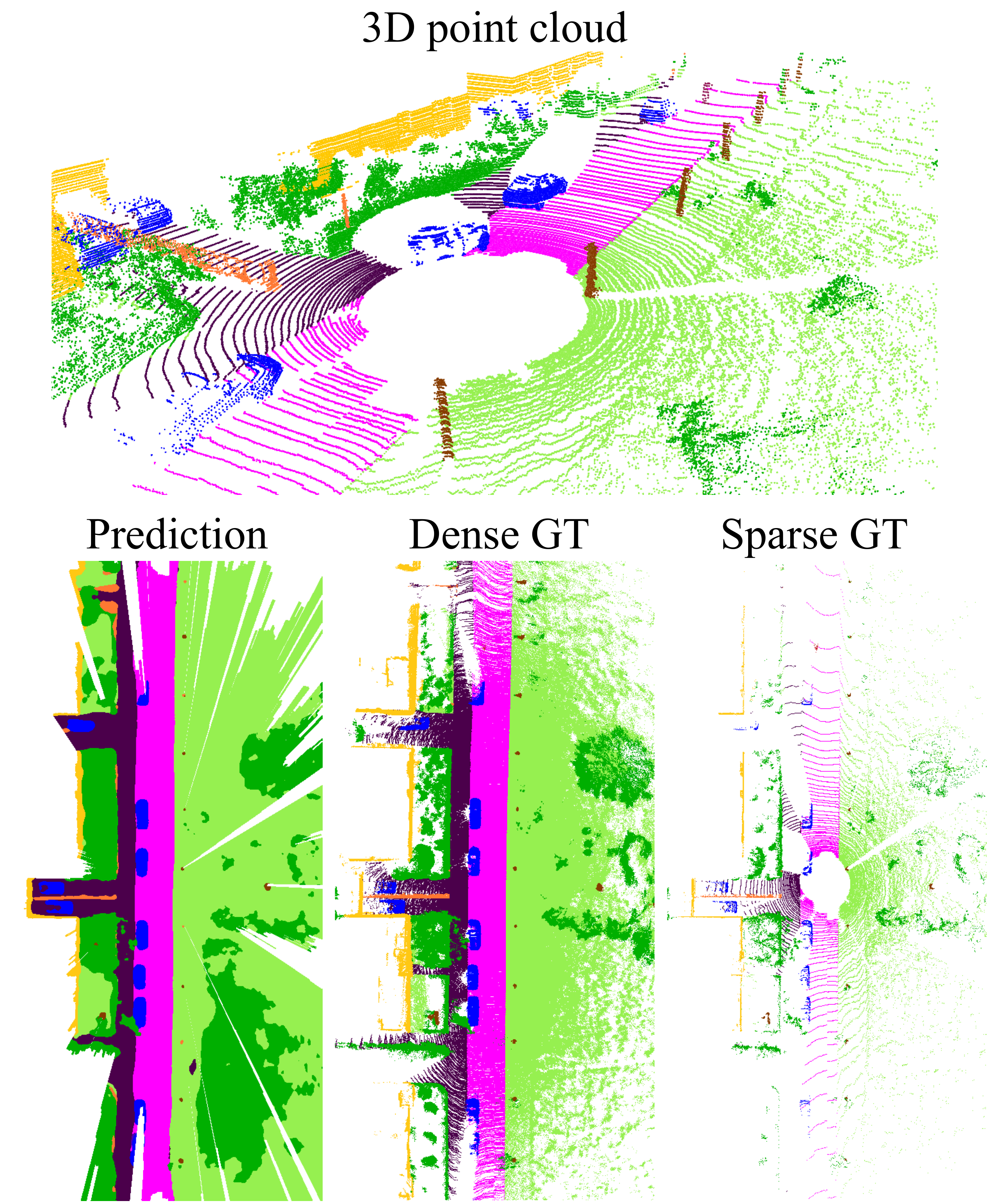}
\end{center}
\caption{
Instead of predicting point-wise semantic scores from the sparse input LiDAR data, our approach outputs a dense top-view semantic grid map (bottom left). 
To enable the training, we generate sparse ground truth (bottom right) from a semantically annotated 3D point cloud (top), and dense ground truth (bottom middle) by superimposing multiple labeled scans.
}
\label{fig:fig1}
\end{figure}

While most existing approaches \cite{milioto2019rangenet, triess2020scan, li2020multi} predict point-wise semantic scores from the sparse LiDAR point cloud, Bieder et al.~\cite{bieder2020exploiting} transform the sparse LiDAR point cloud into a multi-layer grid map representation to obtain a dense top-view segmentation of the LiDAR measurements. In Fig.~\ref{fig:fig1}, we exemplarily show a semantically annotated 3D point cloud and a corresponding dense top-view segmentation result.
We note that such a dense semantic grid map representation is valuable information for subsequent processing in automated vehicles. In particular, 
as an advancement of conventional occupancy grid maps, it can be advantageous to further distinguish free-space areas such as road, sidewalk, or terrain, and occupied areas such as parked vehicles, buildings, or vegetation.

Since the hand-crafted grid map feature extraction in \cite{bieder2020exploiting} can result in a potential information loss, we propose to use \mbox{PointNet} ~\cite{pointnet} to learn features directly from the point cloud and avoid this potential information loss. 
In this paper, we propose a novel end-to-end method named \mbox{PillarSegNet} to approach dense semantic grid map estimation using sparse LiDAR data.
\mbox{PillarSegNet} takes a sparse single sweep LiDAR point cloud as input and subdivides it into a set of pillars in the ground plane. 
The raw point data in the generated pillars is then fed into a simplified \mbox{PointNet} to extract pillar-wise features, which are scattered back into the top view and form a pseudo image. 
This pseudo image is then further consumed by a modified \mbox{U-Net} to predict a dense semantic grid map.

We train and evaluate \mbox{PillarSegNet} on the SemanticKITTI dataset.
To this end, we not only transform the point-wise labels in a single sweep into sparse ground truth represented by 2D grid map, but also accumulate multiple neighbouring labelled scans to obtain dense ground truth.
The dense labels enables a proper evaluation of the dense prediction produced by \mbox{PillarSegNet}.
In addition to the learned pillar features, we further study the benefit of fusing occupancy information obtained via ray-casting. 
Experimental results show that our approach outperforms the state-of-the-art grid map method by a significant margin in term of mIoU.

In summary, the main contributions of our work are:

1)  We propose a novel end-to-end approach to be able to output a semantic grid map given a sparse single LiDAR sweep.

2)  We conduct experiments and quantitative comparisons on the SemanticKITTI dataset to prove the effectiveness of our approach.

3) We perform a comprehensive analysis on the semantic segmentation performance using different feature inputs and data augmentation techniques through an ablation study.

\section{Related Work}
\label{sec:related_work}

\subsection{Point Cloud Representation}
\label{sec:point_cloud_representation}
3D LiDAR point clouds are unstructured, irregular and sparse, making it challenging to process them.
In the domain of automated driving, it is common to represent 3D point clouds in the top view. Such representation has several advantages including scale invariance and minimal occlusions. 
To obtain a top-view representation, there are two main groups of methods: grid mapping and learning-based methods.

\textbf{Grid mapping.} 
Grid mapping approaches, first introduced in \cite{ElfesOccu} are widely used in robotics. The commonly used 2D occupancy grid map encodes the occupancy probability for each evenly spaced grid cell on the ground plane. 
Since the mapping of 3D measurements to 2D implies a loss of information in the height, \cite{hu20wysiwyg} further divides the 3D world into a set of 3D voxels and encodes the occupancy information for each voxel to obtain 3D occupancy grid maps.
In addition to occupancy, other features such as intensity, density and observations can also be derived to form multi-layer grid maps \cite{Wirges2018ObjectDA}. 
In contrast to irregular point sets, the image-like data structure of grid maps enables the use of powerful 
convolutional operations in deep learning. 
Therefore, grid maps are utilized in variety of machine learning applications, e.g. object detection \cite{Wirges2018ObjectDA} and motion estimation \cite{Schreiber2020MotionEI}.
Despite being computationally efficient, grip mapping suffers from an information bottleneck due to the hand-crafted feature extraction, generally leading to a  suboptimal performance. 
For instance, the detection performance for small objects such as pedestrians is noticeably limited in \cite{Wirges2018ObjectDA}.

\textbf{Learning-based.} 
Since the introduction of PointNet \cite{pointnet}, learning-based methods have emerged as prominent alternatives for hand-crafted feature engineering. 
PointNet is an end-to-end neural network that learns point-wise features from the input point set, demonstrating impressive results on indoor 3D object classification and semantic segmentation. 
VoxelNet~\cite{voxelnet} partitions the point cloud into equally spaced voxels and adopts \mbox{PointNet} in novel voxel feature encoding (VFE) layers to learn voxel-wise features. 
The obtained features are then processed with 3D and 2D convolutions for object detection. PointPillars \cite{pointpillars} improves the inference speed by discretizing the point cloud into a set of pillars so that the 3D convolution can be eliminated. 
Pillar features are further used in different perception tasks, e.g. sensor fusion \cite{fei2020semanticvoxels} and ground plane estimation \cite{paigwar2020gndnet}, showing promising results.
In this paper, we exploit pillar features for semantic grid map estimation. 
Although \mbox{PointNet} is able to encode point cloud features very well, we note that occupancy features cannot be learned from the data, as they are obtained via ray-casting and employ an inverse sensor model.
Thus, we optionally use occupancy features to enrich the point cloud features for a better performance. 

\subsection{Image Semantic Segmentation}
Image-based semantic segmentation networks are widely used in autonomous vehicles and significant progress is being made in that field.
As a pioneering work, FCN~\cite{fcn} adopts fully convolutional layers and stacks them in an encoder-decoder fashion with skip connections, enabling the combination of coarse and fine features at different levels. U-Net \cite{unet} builds upon the concept of FCN but uses a larger number of feature channels allowing the propogation of context information to higher resolution layers.
The family of DeepLab~\cite{deeplabv3} uses atrous convolutions and atrous spatial pyramid pooling modules to enlarge \mbox{field-of-views} and improve the contextual understanding. 
In this work, we use a modified U-Net by removing the input convolution block as a trade-off between efficiency and performance.

\subsection{Point Cloud Semantic Segmentation}
For point-wise segmentation of LiDAR data, many approaches map the point cloud onto a top-view image in order to benefit from the progress made in image semantic segmentation. RangeNet++ \cite{milioto2019rangenet} transforms the input point sets into spherical images and uses 2D convolutions for semantic segmentation. Triess et al. \cite{triess2020scan} propose a scan unfolding method and a cyclic padding mechanism to reduce systematic point occlusions during the spherical projection in \cite{milioto2019rangenet}. Instead of using range images, GnDNet \cite{paigwar2020gndnet} encodes features in grid-based representation using PointNet for segmenting the ground points. In a recent work, Bieder et al. \cite{bieder2020exploiting} transform 3D LiDAR data into a multi-layer grid map representation to approach an efficient dense top-view segmentation of point clouds. Nevertheless, it suffers from information loss when generating grid maps and thus performs poor on small objects.

\begin{figure*}
\begin{center}
\includegraphics[width=2.0\columnwidth]{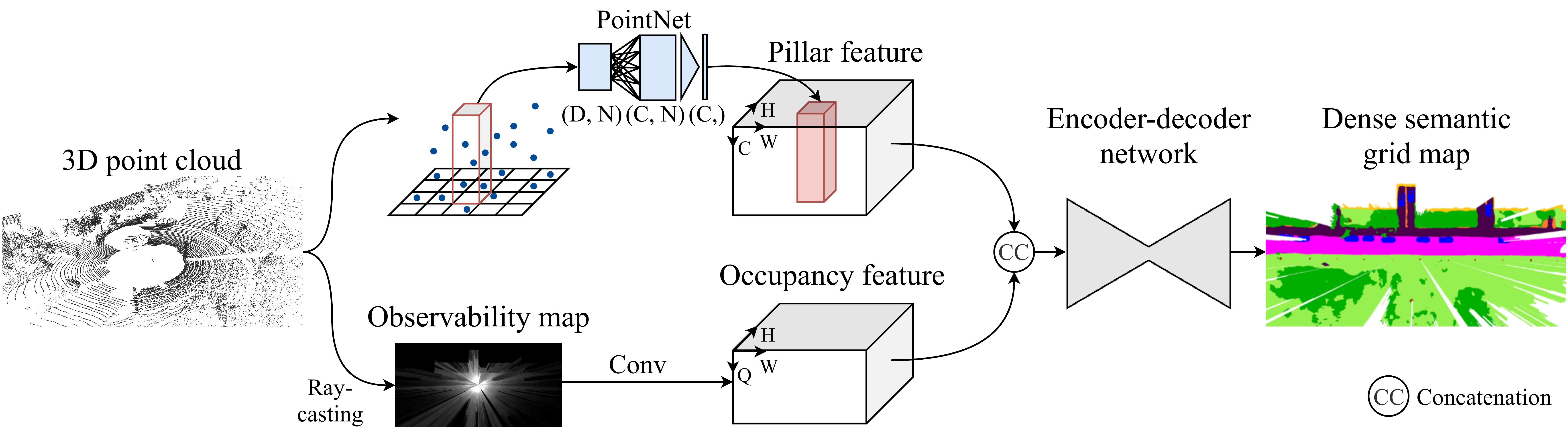}
\end{center}
\caption{
\mbox{PillarSegNet} pipeline overview. 
Given a sparse single-sweep 3D point cloud, \mbox{PillarSegNet} first encodes pillar features and optional occupancy features in two parallel streams. 
The pillar features are encoded using a PointNet, whereas the occupancy features are encoded from an observability map as a result of a model-based ray-casting.
Then, an \mbox{encoder-decoder} network is used to predict a dense semantic grid map from the aggregated features. 
Note that the depicted prediction is filtered by the observability map to exclude occluded areas.
}
\label{fig:pipeline}
\end{figure*}

\section{Method}
\label{sec:method}
An overview of the \mbox{PillarSegNet} pipeline is presented in Fig.~\ref{fig:pipeline}. 
\mbox{PillarSegNet} takes a sparse single-sweep 3D point cloud as input and predicts a dense semantic grid map in the top view. 
It consists of two main blocks, a point cloud feature encoding using parallel pillar and occupancy networks, and a dense semantic segmentation using an encoder-decoder network. 
The details of each block will be presented in the following sections.

\subsection{Point Cloud Feature Encoding}
In the SemanticKITTI dataset, each LiDAR point in the raw point cloud is represented by $(x,y,z,r)$, where $x$, $y$, and $z$ are the 3D coordinates and $r$ indicates the reflectance.
In this work, we consider two kinds of feature encodings obtained from the point cloud: pillar features and occupancy features. 

\textbf{Pillar Features.}
To learn pillar features, the input point cloud is first discretized into a set of pillars in the ground plane, where pillar is a special voxel without an extension bound in the $z$ axis.
For each LiDAR point with $(x,y,z,r)$ encoding in pillars, we follow PointPillars \cite{pointpillars} to further augment it with its offsets from the arithmetic mean $(\Delta{x_c},\Delta{y_c},\Delta{z_c})$ of all points in the pillar and its offsets from the pillar center $(\Delta{x_p},\Delta{y_p},\Delta{z_p})$.
The dimension of the resulting point encoding thus becomes $ D = 10$.
We pre-define the number of pillars per point cloud $(P)$ and the number of points per pillar $(N)$ to create an input tensor of fixed size $(P, N, D)$.
To this end, we randomly sample the points if a pillar has more than $N$ points, or apply zero padding to populate the tensor when a point cloud has less than $P$ pillars or a pillar has too few points.

Each point in pillars is then consumed by a simplified \mbox{PointNet} that consists of a linear layer, BatchNorm and ReLU, outputing a tensor of size $(P, N, C)$.
Then, a max operation along the $N$ axis is applied to create an output tensor of size $(P, C)$, where each pillar-wise feature is represented by a tensor of size $(C)$.
Finally, all pillar-wise features are scattered back to the pillar locations to create a top-view representation of size $(W, H, C)$, where $W$ and $H$ denote the width and height of the grid, corresponding to $x$ and $y$ direction, respectively. 

\textbf{Occupancy Features.}
During the measurement process, the detected LiDAR points are the result of a physical ray-casting. 
When representing the LiDAR points using pillar features, one fundamentally neglects the hidden model information of observability, including information on free space and occupied areas \cite{hu20wysiwyg}.
However, we argue that the observability information might be beneficial for the dense top-view segmentation. 
Hence, we further extract occupancy features and consider to incorporate them in an additional input stream.

In this paper, the grid map framework in \cite{bieder2020exploiting} is adopted to generate a \textit{single-channel} observability map, which indicates the number of transmissions in a grid cell.
The observability map is able to represent occupancy information in the 2D grid.
We further apply a $3 \times 3$ convolutional layer on it to learn 2D occupancy features of size $(W, H, Q)$.
Considering the possible information loss in the height during grid mapping, we also discretize the point cloud into a 3D voxel grid and encode the occupancy probability for each 3D voxel through ray-casting and voxel traversal  \cite{VoxelTraversal}.
In this way, we get a \textit{multi-channel} map which represents 3D occupancy information and use it to replace the observability map.
Similarly, the voxel map is convolved by $3 \times 3$ filters to create 3D occupancy features of size $(W, H, Q)$.

\textbf{Feature Aggregation.}
We adopt the concatenation operation to aggregate pillar features and occupancy features from the two input streams. 
The concatenation operation is simple yet effective for fusing different features~\cite{fei2020semanticvoxels}. 
After concatenation, we obtain aggregated point cloud features of size $(W, H, C + Q)$.

\subsection{Dense Semantic Segmentation}
\label{sec:dense_semantic_segmentation}
To obtain the dense semantic segmentation of the input point cloud features, we use an encoder-decoder network, followed by a segmentation head.

\textbf{Encoder–Decoder Network.}
In this work, we use a modified \mbox{U-Net} which has an encoder-decoder architecture for feature extraction \cite{unet}.
The encoder module gradually halves the spatial size of feature maps and captures higher semantic information, while the decoder module gradually upsamples feature maps and recovers the spatial information.
\mbox{U-Net} also introduces skip connections between the aforementioned modules to combine semantic and spatial information. 
We modify the vanilla \mbox{U-Net} by removing the input convolution block since the aggregated input point cloud features already correspond to a high dimensional feature space. 
This reduces computational and memory overhead with a negligible loss of performance.

\textbf{Segmentation Head.}
After the encoder-decoder network, a set of $1 \times 1$ convolutions is performed in the segmentation head to output logits for the classes in the training data.
During inference, a softmax function is applied over the unbounded logits to provide softmax probabilities for each grid cell.

\textbf{Loss Function.}
Due to the heavy class imbalance in the SemanticKITTI dataset, we adopt the weighted cross entropy loss to optimize the model.
The weights for the classes are defined by considering the class distribution of the ground truth.
The weighted cross entropy loss can be calculated as:
\begin{equation}
    \centering
        \mathcal{L}_{\mathrm{seg}} = -\frac{1}{M}\sum_{i=1}^{M}(\lambda y_i\mathrm{log}\hat{y}_i+(1-\lambda)(1-y_i)\mathrm{log}(1-\hat{y}_i)),
\end{equation}
where $y_i$ and $\hat{y}_i$, respectively, denote the label and softmax probability for pixel $\mathrm{i}$, $\lambda$ is the class-specific weighting coefficient, and $M$ indicates the total number of labeled pixels in the ground truth map.
We choose a coefficient of 2 for class \textit{vehicle} and 8 for classes \textit{pedestrian, two-wheel} and \textit{rider}, when utilizing the sparse ground truth for training.
When the dense ground truth is adopted, the coefficient for class \textit{vehicle} is modified to 5. 
The default coefficients for the remaining classes are set to 1. 
\begin{table*}[t]
\centering
\caption{Quantitative results on the SemanticKITTI validation set.}
\label{tab:experiments_overall}
\begin{tabular}{ll | c | cccccccccccc}
\toprule
Mode
& Method
& \rotatebox{90}{\textbf{mIoU} [\%]}
& \rotatebox{90}{\semcolor[othervehicle] vehicle}
& \rotatebox{90}{\semcolor[person] person}
& \rotatebox{90}{\semcolor[motorcycle] two-wheel}
& \rotatebox{90}{\semcolor[motorcyclist] rider}
& \rotatebox{90}{\semcolor[road] road}
& \rotatebox{90}{\semcolor[sidewalk] sidewalk}
& \rotatebox{90}{\semcolor[parking] other-ground}
& \rotatebox{90}{\semcolor[building] building}
& \rotatebox{90}{\semcolor[fence] object}
& \rotatebox{90}{\semcolor[vegetation] vegetation}
& \rotatebox{90}{\semcolor[trunk] trunk}
& \rotatebox{90}{\semcolor[terrain] terrain} \\
\midrule
\multirow{4}{*}{\shortstack[l]{Sparse Train\\Sparse Eval}}
& Bieder et al. \cite{bieder2020exploiting}           & \textbf{39.8} & 69.7 & 0.0  & 0.0  & 0.0  & 85.8 & 60.3 & 25.9 & 72.8 & 15.1 & 68.9 & 9.9  & 69.3 \\
& Pillar                 & \textbf{55.1} & 79.5 & 15.8 & 25.8 & 51.8 & 89.5 & 70.0 & 38.9 & 80.6 & 25.5 & 72.8 & 38.1 & 72.7  \\
& Pillar + 2D Occupancy  & \textbf{55.3} & 82.7 & 20.3 & 24.5 & 51.3 & 90.0 & 71.2 & 36.5 & 81.3 & 28.3 & 70.4 & 38.5 & 69.0 \\
& Pillar + 3D Occupancy  & \textbf{56.2} & 83.8 & 19.5 & 24.8 & 51.8 & 90.1 & 72.3 & 36.9 & 81.5 & 28.2 & 72.1 & 41.7 & 71.5   \\
\midrule
\multirow{4}{*}{\shortstack[l]{Sparse Train\\Dense Eval}}
& Bieder et al. \cite{bieder2020exploiting}          & \textbf{32.8} & 43.3 & 0.0 & 0.0 & 0.0 & 84.3 & 51.4 & 22.9 & 54.7 & 10.8 & 51.0 & 6.3 & 68.6  \\
& Pillar                 & \textbf{37.5} & 45.1 & 0.0 & 0.1 & 3.3 & 82.7 & 57.5 & 29.7 & 64.6 & 14.0 & 58.5 & 25.5 & 68.9   \\
& Pillar + 2D Occupancy  & \textbf{38.4} & 52.5 & 0.0 & 0.2 & 3.0 & 85.6 & 60.1 & 29.8 & 65.7 & 16.1 & 56.7 & 26.2 & 64.5   \\
& Pillar + 3D Occupancy  & \textbf{38.9} & 53.3 & 0.0 & 0.1 & 5.0 & 86.1 & 60.5 & 29.8 & 65.1 & 16.4 & 56.7 & 28.1 & 66.2   \\
\midrule
\multirow{3}{*}{\shortstack[l]{Dense Train\\Dense Eval}}
& Pillar                 & \textbf{42.8} & 70.3 & 5.4 & 6.0 & 8.0  & 89.8 & 65.7 & 34.0 & 65.9 & 16.3 & 61.2 & 23.5 & 67.9   \\
& Pillar + 2D Occupancy  & \textbf{44.1} & 72.8 & 7.4 & 4.7 & 10.2 & 90.1 & 66.2 & 32.4 & 67.8 & 17.4 & 63.1 & 27.6 & 69.2   \\
& Pillar + 3D Occupancy  & \textbf{44.6} & 73.1 & 7.8 & 6.0 & 10.0 & 89.7 & 65.7 & 30.3 & 68.3 & 18.4 & 65.0 & 30.4 & 70.8   \\
\bottomrule
\end{tabular}
\end{table*}

\begin{figure*}[t]
\begin{center}
\includegraphics[width = 1.0\textwidth]{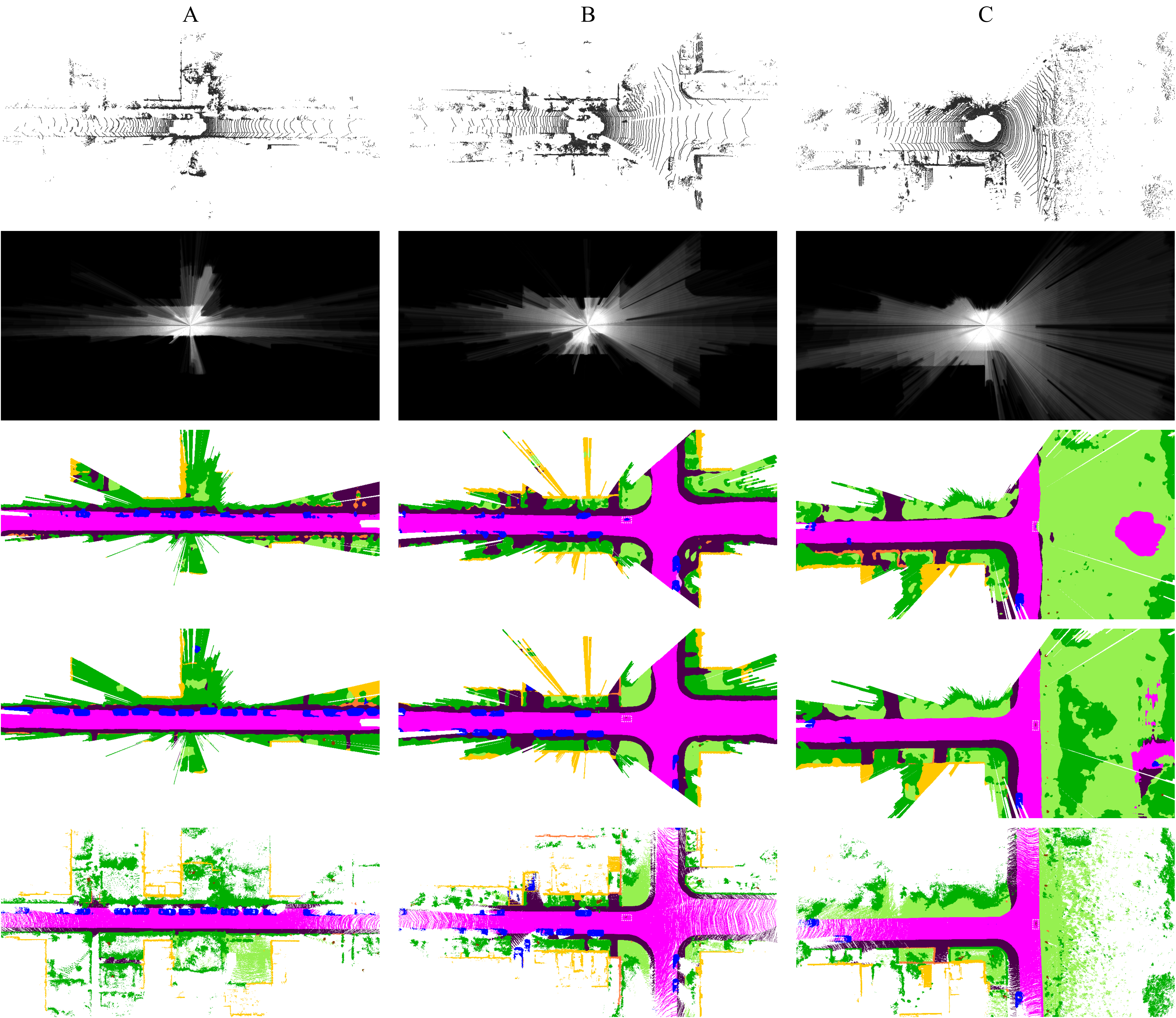}
\end{center}
\caption{Qualitative results produced by Bieder et al. \cite{bieder2020exploiting} and our approach on the SemanticKITTI validation set. 
Our network is solely based on the \textit{Pillar} input features and was trained on the dense ground truth. 
From top to bottom in each column, we depict the input point cloud, the observability map, the prediction from  \cite{bieder2020exploiting}, the prediction from our approach, as well as the corresponding ground truth.
The unobservable areas in each prediction map were filtered out according to the observability map.
In comparison with \cite{bieder2020exploiting}, our approach shows more accurate prediction on vehicles (cases A \& B) and small objects, e.g. two-wheels (cases B \& C),
Best viewed digitally with zoom.}
\label{fig:qualitative}
\end{figure*}  

\section{Training and Experiments}
\label{sec:training_experiments}
In this section, we introduce the dataset used to train and validate our approach together with the experimental setup.

\subsection{Dataset}
\label{sec:dataset}
We evaluate our approach on the SemanticKITTI dataset \cite{behley2019iccv, kitti}, which is based on a subset of the KITTI odometry dataset sequences, and contains accurate pose information and semantic labels for each LiDAR point.
As in \cite{behley2019iccv, bieder2020exploiting}, we use sequences 00-07 and 09-10 for training and sequence 08 for validation.
In this case, the training and validation sets contain 19130 and 4071 LiDAR scans, respectively.
To tackle the under-representation of rare classes, we apply the same merging of 19 into 12 classes as in \cite{bieder2020exploiting}, which also enables a fair comparison.
In particular, we map the classes \textit{car, truck} and \textit{other-vehicle} to \textit{vehicle}, the classes \textit{motorcyclist} and \textit{bicyclist} to \textit{rider}, the classes \textit{bicycle} and \textit{motorcycle} to \textit{two-wheel}, the classes \textit{traffic-sign}, \textit{pole} and \textit{fence} to \textit{object}, and the classes \textit{other-ground} and \textit{parking} to \textit{other-ground}. 
All \textit{unlabeled} pixels are not considered when optimizing the model.

\subsection{Ground Truth Generation}
\textbf{Sparse Ground Truth Generation.}
\label{sec:sparse_gt}
To obtain the top-view sparse ground truth, we first rasterize a point-wise labelled point cloud into grid cells. 
For each grid cell, we count the number of points for each of the $12$ classes. 
The semantic class $k_i$ for grid cell $i$ is then determined through a weighted $\mathrm{argmax}$ operation:
\begin{equation}
\label{eq:argmax}
\centering
k_{i} = \underset{k\in[1,K]}{\mathrm{argmax}} \left( w_k  n_{i,k}\right),
\end{equation}
where $K$ is the number of classes, $n_{i,k}$ denotes the number of points of class $k$ inside grid cell $i$, and $w_k$ is the weighting factor for class $k$.
For traffic participant classes including \textit{vehicle}, \textit{person}, \textit{rider} and \textit{two-wheel}, we choose a factor of 5.
To ignore the class \textit{unlabeled}, we set the corresponding factor to 0.
For all remaining classes, we use a factor of 1.
Grid cells without any assigned points are labeled as \textit{unlabeled}. An example of a sparse ground truth map is depicted in Fig.~\ref{fig:fig1}.

\textbf{Dense Ground Truth Generation}
\mbox{SemanticKITTI} contains consecutive LiDAR scans with accurate pose information, which allows the generation of dense ground truth by
superimposing multiple labeled scans.
For each scan of a sequence, we first collect a set of neighbouring point clouds with a sensor distance $|\Delta p_x|$ smaller than twice the farthest point distance $d$, while limiting the maximal amount of scans to 40. 
With the provided poses, the selected point clouds are then transformed to the LiDAR coordinate system of the current scan.
To avoid aliasing of moving objects during aggregation, we only aggregate points that belong to static objects.
For moving objects, we only use information from the current scan.
We follow the process used for sparse ground truth generation to create dense ground truth maps.
An example of a dense ground truth map is shown in Fig.~\ref{fig:fig1}.

\subsection{Implementation Details}
\mbox{PillarSegNet} takes a sparse single LiDAR sweep as input and outputs semantic probabilities for the $12$ classes described in Section~\ref{sec:dataset}.
We implement \mbox{PillarSegNet} based on the OpenPCDet codebase\footnote{\url{https://github.com/open-mmlab/OpenPCDet}}.
For all experiments, we crop the input point cloud at $[(-50, 50), (-25, 25), (-2.5, 1.5)]$ meters along $x$, $y$, $z$ axes respectively.
When generating pillars, we use a pillar grid size of $0.1^2\,\mathrm{m}^2$, number of pillars $P=30000$, and number of points per pillar $N=20$. The channels of the pillar-wise feature learned by PointNet is defined as $C=64$. The grid cell for generating the observability map and the voxel for encoding the 3D occupancy map have the same $xy$ resolution as the pillar, while the voxel has a $z$ resolution of $0.2$ meters.
The occupancy features after convolutions have $Q=16$ channels.

During training, multiple data augmentation techniques, namely \textit{random flip, random rotation, random scaling} and \textit{random translation}, are applied to prevent the network from overfitting on the training data. 
In \textit{random flip}, every LiDAR point is flipped along the $x$ or $y$ axis by a 50\% chance.
\textit{Random rotation} means that we rotate every point around the upright $z$ axis by an uniformly distributed angle between $(-\frac{\pi}{4}, +\frac{\pi}{4})$ radians.
In \textit{random scaling}, every point is scaled by a scalar sampled from a uniform distribution between (0.95, 1.05).
\textit{Random translation} means that every point is shifted by $(\Delta{x},\Delta{y},\Delta{z})$, where $\Delta{x},\Delta{y}$ and $\Delta{z}$ are sampled independently from normal distributions \mbox{\textit{N} (0, $\sigma^2$)}. We set $\sigma$ as (5, 5, 0.05) meters for $(\Delta{x},\Delta{y},\Delta{z})$ respectively.
When taking occupancy features as input, we apply respective data augmentation techniques on the input occupancy map to make pillar and occupancy features consistent. 

The network is trained from scratch using the Adam optimizer~\cite{adam} with an initial learning rate of 0.001, and weight decay of 0.01. The training lasts 30 epochs with a mini-batch size of 2.

\section{Results}
\label{sec:results}
\subsection{Quantitative Analysis}
\textbf{Metrics.}
We use the commonly applied Intersection over Union (IoU) \cite{pascal}, or Jaccard index, to evaluate the performance quantitatively. 
The IoU for class $k$ is calculated by
\begin{equation}
    \centering
        \mathrm{IoU}_k = \frac{\mathrm{TP}_k}{\mathrm{TP}_k+\mathrm{FP}_k+\mathrm{FN}_k},
\end{equation}
where $\mathrm{TP}_k, \mathrm{FP}_k$, and $\mathrm{FN}_k$, respectively, correspond to the number of true positive, false positive, and false negative predictions. 

To assess the overall performance, we compute the mean IoU (mIoU) by
\begin{equation}
    \centering
        \mathrm{mIoU} = \frac{1}{K}\sum_{k=1}^{K}{\mathrm{IoU}_k}.
\end{equation}

\textbf{Experimental Results.}
We train the network in two modes, namely \textit{Sparse Train} and \textit{Dense Train}, according to which ground truth is used.
The former considers the sparse ground truth, which is obtained from single sweep, whereas the latter considers the dense ground truth, which is obtained by aggregating multiple sweeps.
For each experiment using \textit{Sparse Train} mode, we use two approaches for evaluation: \textit{Sparse Eval} and \textit{Dense Eval}, in order to achieve a fair comparisons with the method in \cite{bieder2020exploiting}.
The former considers the sparse ground truth for metrics calculation, whereas the latter considers the dense ground truth, and additionally filters the predictions with the observability map to exclude occluded areas. 
Note that we only evaluate the \textit{Dense Train} experiments by the \textit{Dense Eval} approach.

The experimental results on the full validation set are presented in Table~\ref{tab:experiments_overall}.
We first focus on the results obtained in the \textit{Sparse Train} mode.
Our pillar-based method outperforms the state-of-the-art grid map method \cite{bieder2020exploiting} by 15.3\% and 5.7\% mIoU in \textit{Sparse Eval} and \textit{Dense Eval}, respectively.
In particular, our method achieves a large IoU improvement in the \textit{Sparse Eval} mode for small object classes, such as {\it person}, {\it two-wheel}, and {\it rider}. 
These significant performance gains indicate the superiority of the learned pillar features over the grid map representation.

In addition to pillar features, we further analyse the role of 2D and 3D occupancy features in semantic grid map estimation. 
By aggregating respective occupancy features, our method achieves an additional improvement in terms of mIoU performance, where the 3D variant performs slightly better when compared with the 2D variant.
This demonstrates that the observability information, generated via model-based ray-casting, can be successfully used to further improve the performance of semantic grid map estimation. 

We now consider the results obtained in the \textit{Dense Train} mode. When compared with the \textit{Sparse Train} mode, training the network on the dense ground truth leads to notable IoU improvements for all classes.
For the deployment of semantic grid map estimation, it is thus advisable to leverage dense labels for better performance. 
Consistent with our previous observation, 2D and 3D occupancy information yield close overall performance gains.
Considering the increased computational overhead when generating the 3D occupancy map, it is more practical to aggregate 2D occupancy features and pillar features as a compromise between performance and computational efficiency.

\textbf{Inference Time.}
We measure the inference time of \mbox{PillarSegNet} on a desktop computer with an Intel i9 CPU and a Nvidia 2080 Ti GPU.
Our pillar-based approach achieves a total runtime of \mbox{58\,ms}. When incorporating 2D occupancy features, additional \mbox{10\,ms} and \mbox{6\,ms} are needed for grid mapping and in the network, respectively.

\textbf{Ablation Study.}
We perform thorough ablation experiments to investigate the effect of different data augmentation techniques in our method. 
The results are reported in Table~\ref{tab:experiments_da}.
Baseline denotes the method that takes pillar features as input and is trained on the sparse ground truth without any data augmentations.
All experiments are evaluated in the \textit{Sparse Eval} mode.
We observe that \textit{random flip} and \textit{random rotation} significantly boost the performance with 4.6\% gain in terms of mIoU, while \textit{random scale} contributes to another 0.6\% improvement.
Unexpectedly, \textit{random translation} has a negative effect on the overall performance, presumably, because it violates the condition that the ego car should always be in the grid center.

\begin{table}[h]
\centering
\caption{Ablation study for data augmentation techniques on the SemanticKITTI validation set.}
\label{tab:experiments_da}
\begin{tabular}{cccccc}
\toprule
Baseline & Flip & Rotate & Scale & Translate & mIoU [\%] \\
\midrule
\checkmark & & & & & 50.4 \\
\checkmark & \checkmark & & & & 53.0 \\
\checkmark & \checkmark & \checkmark & & & 55.0 \\
\checkmark & \checkmark & \checkmark & \checkmark & & 55.6 \\
\checkmark & \checkmark & \checkmark & \checkmark & \checkmark & 55.1 \\
\bottomrule
\end{tabular}
\end{table}

\subsection{Qualitative Analysis}
\label{sec:qualitative}

In Fig.~\ref{fig:qualitative}, we show qualitative results obtained from the approach proposed by Bieder et al. \cite{bieder2020exploiting} as well as our approach which is based on the \mbox{\textit{Pillar}} features and was trained on the dense labels.
In comparison with the dense ground truth, our approach accurately segments general road scenes, which include the static environment and varying traffic participants. 
This demonstrates the ability of our approach for semantic scene understanding from a sparse single LiDAR scan.
For instance, when compared with the grid map method, our approach shows more accurate prediction for vehicles (cases A \& B) or small objects (cases B \& C), indicating the superiority of the learning-based feature extraction.

A video showing qualitative results on the full validation set obtained by our approach is available at:\\ 
\href{https://youtu.be/3yidrXBnfo4}{https://youtu.be/3yidrXBnfo4}.

\section{Conclusion}
\label{sec:conclusion}
In this work, we presented a novel end-to-end approach named \mbox{PillarSegNet} for semantic grid map estimation based on sparse LiDAR data.
\mbox{PillarSegNet} learns features directly from the 3D point cloud, aiming to avoid the information bottleneck caused by hand-crafted feature engineering.
Experimental results on the SemanticKITTI dataset show that \mbox{PillarSegNet} outperforms the state-of-the-art grid map method by a large margin in terms of mIoU. 
We further incorporate occupancy information obtained via \mbox{ray-casting} to enhance the overall performance and identify the 2D occupancy information as the best trade-off between performance and computational efficiency.

\section*{Acknowledgment}
\label{sec:acknowledgment}
The research leading to these results is funded by the German Federal Ministry for Economic Affairs and Energy within the project “Methoden und Maßnahmen zur Absicherung von KI basierten Wahrnehmungsfunktionen für das automatisierte Fahren (KI-Absicherung)". The authors would like to thank the consortium for the successful cooperation.

\bibliographystyle{IEEEtran}
\bibliography{root}

\begin{thebibliography}{10}
\providecommand{\url}[1]{#1}
\csname url@samestyle\endcsname
\providecommand{\newblock}{\relax}
\providecommand{\bibinfo}[2]{#2}
\providecommand{\BIBentrySTDinterwordspacing}{\spaceskip=0pt\relax}
\providecommand{\BIBentryALTinterwordstretchfactor}{4}
\providecommand{\BIBentryALTinterwordspacing}{\spaceskip=\fontdimen2\font plus
\BIBentryALTinterwordstretchfactor\fontdimen3\font minus
  \fontdimen4\font\relax}
\providecommand{\BIBforeignlanguage}[2]{{%
\expandafter\ifx\csname l@#1\endcsname\relax
\typeout{** WARNING: IEEEtran.bst: No hyphenation pattern has been}%
\typeout{** loaded for the language `#1'. Using the pattern for}%
\typeout{** the default language instead.}%
\else
\language=\csname l@#1\endcsname
\fi
#2}}
\providecommand{\BIBdecl}{\relax}
\BIBdecl

\bibitem{behley2019iccv}
J.~Behley, M.~Garbade, A.~Milioto, J.~Quenzel, S.~Behnke, C.~Stachniss, and
  J.~Gall, ``{SemanticKITTI: A Dataset for Semantic Scene Understanding of
  LiDAR Sequences},'' in \emph{IEEE/CVF International Conference on Computer
  Vision (ICCV)}, 2019, pp. 9296--9306.

\bibitem{milioto2019rangenet}
A.~Milioto, I.~Vizzo, J.~Behley, and C.~Stachniss, ``{RangeNet++: Fast and
  Accurate LiDAR Semantic Segmentation},'' in \emph{IEEE/RSJ International
  Conference on Intelligent Robots and Systems (IROS)}, 2019, pp. 4213--4220.

\bibitem{triess2020scan}
L.~T. Triess, D.~Peter, C.~B. Rist, and J.~M. Z{\"o}llner, ``{Scan-based
  Semantic Segmentation of LiDAR Point Clouds: An Experimental Study},'' in
  \emph{IEEE Intelligent Vehicles Symposium (IV)}, 2020, pp. 1116--1121.

\bibitem{li2020multi}
S.~Li, X.~Chen, Y.~Liu, D.~Dai, C.~Stachniss, and J.~Gall, ``{Multi-scale
  Interaction for Real-time LiDAR Data Segmentation on an Embedded Platform},''
  \emph{arXiv preprint arXiv:2008.09162}, 2020.

\bibitem{bieder2020exploiting}
F.~Bieder, S.~Wirges, J.~Janosovits, S.~Richter, Z.~Wang, and C.~Stiller,
  ``{Exploiting Multi-Layer Grid Maps for Surround-View Semantic Segmentation
  of Sparse LiDAR Data},'' in \emph{IEEE Intelligent Vehicles Symposium (IV)},
  2020, pp. 1892--1898.

\bibitem{pointnet}
C.~R. Qi, H.~Su, K.~Mo, and L.~J. Guibas, ``{PointNet: Deep Learning on Point
  Sets for 3D Classification and Segmentation},'' in \emph{IEEE Conference on
  Computer Vision and Pattern Recognition (CVPR)}, 2017, pp. 652--660.

\bibitem{ElfesOccu}
A.~Elfes, ``{Using Occupancy Grids for Mobile Robot Perception and
  Navigation},'' \emph{Computer}, vol.~22, no.~6, pp. 46--57, Jun. 1989.

\bibitem{hu20wysiwyg}
P.~Hu, J.~Ziglar, D.~Held, and D.~Ramanan, ``{What You See Is What You Get:
  Exploiting Visibility for 3d Object Detection},'' in \emph{IEEE/CVF
  Conference on Computer Vision and Pattern Recognition (CVPR)}, 2020, pp.
  11\,001--11\,009.

\bibitem{Wirges2018ObjectDA}
S.~Wirges, T.~Fischer, J.~Frias, and C.~Stiller, ``{Object Detection and
  Classification in Occupancy Grid Maps Using Deep Convolutional Networks},''
  in \emph{IEEE International Conference on Intelligent Transportation Systems
  (ITSC)}, 2018, pp. 3530--3535.

\bibitem{Schreiber2020MotionEI}
M.~Schreiber, V.~Belagiannis, C.~Gl{\"a}ser, and K.~Dietmayer, ``{Motion
  Estimation in Occupancy Grid Maps in Stationary Settings Using Recurrent
  Neural Networks},'' in \emph{IEEE International Conference on Robotics and
  Automation (ICRA)}, 2020, pp. 8587--8593.

\bibitem{voxelnet}
Y.~Zhou and O.~Tuzel, ``{VoxelNet: End-to-End Learning for Point Cloud Based 3D
  Object Detection},'' in \emph{IEEE Conference on Computer Vision and Pattern
  Recognition (CVPR)}, 2018, pp. 4490--4499.

\bibitem{pointpillars}
A.~H. Lang, S.~Vora, H.~Caesar, L.~Zhou, J.~Yang, and O.~Beijbom,
  ``{{PointPillars}: Fast Encoders for Object Detection From Point Clouds},''
  in \emph{IEEE Conference on Computer Vision and Pattern Recognition (CVPR)},
  2019, pp. 12\,697--12\,705.

\bibitem{fei2020semanticvoxels}
J.~Fei, W.~Chen, P.~Heidenreich, S.~Wirges, and C.~Stiller, ``{SemanticVoxels:
  Sequential Fusion for 3D Pedestrian Detection using LiDAR Point Cloud and
  Semantic Segmentation},'' in \emph{IEEE International Conference on
  Multisensor Fusion and Integration for Intelligent Systems (MFI)}, 2020, pp.
  185--190.

\bibitem{paigwar2020gndnet}
A.~Paigwar, {\"O}.~Erkent, D.~S. Gonz{\'a}lez, and C.~Laugier, ``{GndNet: Fast
  Ground Plane Estimation and Point Cloud Segmentation for Autonomous
  Vehicles},'' in \emph{IEEE/RSJ International Conference on Intelligent Robots
  and Systems (IROS)}, 2020, pp. 2150--2156.

\bibitem{fcn}
J.~Long, E.~Shelhamer, and T.~Darrell, ``{Fully Convolutional Networks for
  Semantic Segmentation},'' in \emph{IEEE Conference on Computer Vision and
  Pattern Recognition (CVPR)}, 2015, pp. 3431--3440.

\bibitem{unet}
O.~Ronneberger, P.~Fischer, and T.~Brox, ``{U-net: Convolutional networks for
  biomedical image segmentation},'' in \emph{International Conference on
  Medical Image Computing and Computer Assisted Intervention}.\hskip 1em plus
  0.5em minus 0.4em\relax Springer, 2015, pp. 234--241.

\bibitem{deeplabv3}
L.-C. Chen, Y.~Zhu, G.~Papandreou, F.~Schroff, and H.~Adam, ``{Encoder-Decoder
  With Atrous Separable Convolution for Semantic Image Segmentation},'' in
  \emph{European Conference on Computer Vision (ECCV)}, 2018, pp. 801--818.

\bibitem{VoxelTraversal}
J.~Amanatides and A.~Woo, ``{A Fast Voxel Traversal Algorithm for Ray
  Tracing},'' in \emph{EG 1987-Technical Papers}.\hskip 1em plus 0.5em minus
  0.4em\relax Eurographics Association, 1987.

\bibitem{kitti}
A.~Geiger, P.~Lenz, and R.~Urtasun, ``{Are we ready for Autonomous Driving? The
  KITTI Vision Benchmark Suite},'' in \emph{IEEE Conference on Computer Vision
  and Pattern Recognition (CVPR)}, 2012, pp. 3354--3361.

\bibitem{adam}
P.~D. Kingma and L.~J. Ba, ``{Adam: A Method for Stochastic Optimization},'' in
  \emph{International Conference on Learning Representations (ICLR)}, 2015.

\bibitem{pascal}
M.~Everingham, S.~A. Eslami, L.~Van~Gool, C.~K. Williams, J.~Winn, and
  A.~Zisserman, ``{The PASCAL Visual Object Classes Challenge: A
  Retrospective},'' \emph{International Journal of Computer Vision}, vol. 111,
  no.~1, pp. 98--136, 2015.

\end{thebibliography}

\end{document}